# Adversarial example generation with AdaBelief Optimizer and Crop Invariance

Bo Yang, Hengwei Zhang, Yuchen Zhang, Kaiyong Xu, Jindong Wang

## Abstract

Deep neural networks are vulnerable to adversarial examples, which are crafted by applying small, human-imperceptible perturbations on the original images, so as to mislead deep neural networks to output inaccurate predictions. Adversarial attacks can thus be an important method to evaluate and select robust models in safety-critical applications. However, under the challenging black-box setting, most existing adversarial attacks often achieve relatively low success rates on adversarially trained networks and advanced defense models. In this paper, we propose AdaBelief Iterative Fast Gradient Method (ABI-FGM) and Crop-Invariant attack Method (CIM) to improves the transferability of adversarial examples. ABI-FGM and CIM can be readily integrated to build a strong gradient-based attack to further boost the success rates of adversarial examples for black-box attacks. Moreover, our method can also be naturally combined with other gradient-based attack methods to build a more robust attack to generate more transferable adversarial examples against the defense models. Extensive experiments on the ImageNet dataset demonstrate the method's effectiveness. Whether on adversarially trained networks or advanced defense models, our method has higher success rates than state-of-the-art gradient-based attack methods.

## 1. Introduction

In the field of image recognition, deep neural networks are capable of classifying images with performance superior to that of humans (Krizhevsky et al., 2017; Simonyan et al., 2014; Szegedy et al, 2015; He et al., 2016). However, researchers have found that deep neural networks are very fragile. Szegedy et al. (2015) found for the first time the intriguing properties of deep neural networks: adding small, human-imperceptible perturbations to the original image, can make deep neural networks give an erroneous output with high confidence. These disturbed images are adversarial examples, in addition, adversarial examples with strong attack performance are an important tool to evaluate the robustness of the model. Studying how to generate more transferable adversarial examples can help to assess and improve the robustness of the models.

With the knowledge of the network structure and weights, many methods can successfully generate adversarial examples and perform white-box attacks, including optimization-based methods such as box-constrained L-BFGS (Szegedy et al., 2015), one-step gradient-based method such as fast gradient sign method (Goodfellow et al., 2014), multi-step iterative gradient-based methods such as iterative fast gradient sign method (Alexey et al., 2016) and momentum iterative fast gradient sign method (Dong et al., 2018), and Carlini & Wagner attack method (Carlini et al., 2017). However, under the white-box setting, the attackers need to fully know the structure and parameters of a given model, which is difficult to achieve in the adversarial environment. In general, adversarial examples generated for one model may also be adversarial to others, i.e., adversarial examples have a certain degree of transferability, which enables black-box attacks and poses real security issues.

Despite adversarial examples are generally transferable, to further improve their transferability for effective black-box attacks remains to be explored. Xie et al (2019) propose a diverse input method based on data augmentation to improve the transferability of adversarial examples. Dong et al (2019) propose a translation-invariant attack method to generate adversarial examples that are less sensitive to the discriminative regions of the white-box model being attacked, and have higher transferability against the defense models. Lin et al (2020) propose a scale-invariant attack method to generate more transferable adversarial examples. However, these existing methods often exhibit low success rates under the black-box setting, especially for defense models and adversarially trained networks.

To this end, by regarding the adversarial example generation process as an optimization process similar to deep neural network training, we propose two new methods to generate more transferable adversarial examples, namely AdaBelief Iterative Fast Gradient Method (ABI-FGM) and Crop-Invariant attack Method (CIM).

- We introduce Adabelief optimizer into iterative gradient attack to form AdaBelief Iterative Fast Gradient Method. This method adopts adaptive learning rate, and considers the difference between the first moment and gradient to control the size of forward step in the iterative process, which effectively optimizes the convergence process and improves the transferability of adversarial examples.

- Inspired by data enhancement (Krizhevsky et al., 2017; Simonyan et al., 2014), we find that image edge information has little influence on the correct classification of images, and crop the image edge within a certain range can reduce the overfitting, so we find that deep neural networks have crop-invariant property. Based on this, we propose a Crop-Invariant attack Method, which optimizes the adversarial perturbations over the crop copies of the input images to generate more transferable adversarial examples.

- Besides, ABI-FGM and CIM can be naturally integrated to build a strong gradient-based attack to further boost the success rates of adversarial examples for black-box attacks. Moreover, combining our ABI-FGM and CIM with existing gradient-based attack methods (e.g., diverse input method (Xie et al., 2019), translation-invariant method (Dong et al., 2019)) can further boost the attack success rates of adversarial examples.

Extensive experiments on the ImageNet dataset (Russakovsky et al., 2015) show that our method have higher success rates of black-box attack in both normally trained models and adversarially trained models than existing baseline attack methods. In addition, by integrating with baseline attacks, the attack success rates of our method under the black-box setting are further improved. In particular, our best method, CI-AB-SI-TI-DIM (Crop-Invariant AdaBelief Iterative FGM integrated with scale-invariant and translation-invariant diverse input method), achieves an average success rate of 97.3% against seven models under multi-model setting. We hope that the proposed attack method can help evaluate the robustness of models and effectiveness of defense methods.

## 2. Related Work

In this section, we first provide the background knowledge, and then briefly sort out the related works about adversarial attack and defense methods. Let $x$ and $y$ be an input image and the corresponding ground-truth label, respectively. The term $\theta$ represents neural network parameters and $J(\theta, x, y)$ denotes a loss function, usually the cross-entropy loss. The primary objective is to mislead the model by maximizing $J(\theta, x, y)$ to generate a visually indistinguishable adversarial example $x^{adv}$ from $x$, that is, to make the model misclassified. We used an $L_\infty$ norm bound in this work to measure the distortion between $x$ and $x^{adv}$, such that $\|x^* - x\|_\infty \leq \varepsilon$, where $\varepsilon$ is the magnitude of the perturbation. Therefore, the generation of adversarial examples can be converted into solving the following constrained optimization problem:

$$\arg\max_{x^{adv}} J(\theta, x^{adv}, y), \quad s.t. \|x^{adv} - x\|_\infty \leq \varepsilon. \quad (1)$$

### 2.1. Attack Method

Here we provide a brief introduction of several attack methods of generating adversarial examples.

**Fast Gradient Sign Method (FGSM).** FGSM (Goodfellow et al., 2014) is one of the most basic methods to generate adversarial examples, which seeks the adversarial perturbations in the direction of the loss gradient. The method can be expressed as

$$x^{adv} = x + \varepsilon \cdot sign(\nabla_x J(\theta, x, y)), \quad (2)$$

Where $sign(\cdot)$ is the sign function to limit the distortion between $x$ and $x^{adv}$ in the $L_\infty$ norm bound.

**Iterative Fast Gradient Sign Method (I-FGSM).** I-FGSM (Alexey et al., 2016) is an iterative version of FGSM, which iteratively apply fast gradient multiple times with a small step size $\alpha$. The update equation is as follows:

$$x_{t+1}^{adv} = \text{Clip}_x^\varepsilon \{x_t^{adv} + \alpha \cdot sign(\nabla_x J(\theta, x_t^{adv}, y))\}, \quad (3)$$

Where $\alpha = \varepsilon / T$, in which $T$ is the number of iterations. The function $\text{Clip}_x^\varepsilon(\cdot)$ make the generated adversarial examples satisfy the $L_\infty$ norm bound. It has been shown that I-FGSM has a higher success rate of white-box attack than FGSM at the cost of worse transferability.

**Momentum Iterative Fast Gradient Sign Method (MI-FGSM).** MI-FGSM (Dong et al., 2018) proposes to integrate the momentum term into iterative attack method to stabilize update directions, so as to improve the transferability of adversarial examples. The update procedure is as follows:

$$g_0 = 0, \quad g_{t+1} = \mu \cdot g_t + \frac{\nabla_x J(\theta, x_t^{adv}, y)}{\|\nabla_x J(\theta, x_t^{adv}, y)\|_1}, \quad (4)$$

$$x_{t+1}^{adv} = \text{Clip}_x^\varepsilon \{x_t^{adv} + \alpha \cdot sign(g_{t+1})\}, \quad (5)$$

Where $g_t$ is the accumulated gradient at iteration $t$ and $\mu$ is the decay factor of the momentum term.

**Diverse Input Method (DIM).** DIM (Xie et al., 2019) applies random transformations to the original inputs with a given probability at each iteration to alleviate the overfitting phenomenon. The transformations include the random resizing and the random padding. This method can be integrated into other baseline attack methods to further improve the transferability.

**Translation-Invariant Method (TIM).** Dong et al. (2019) proposes a translation-invariant attack method to generate adversarial examples that are less sensitive to the discriminative regions of the white-box model being attacked, and have higher transferability against the defense models. To improve the efficiency of attacks, they further implement their method by convolving the gradient at the untranslated image with a pre-defined kernel matrix. TIM can also be combined with other gradient-based attack methods to generate more transferable adversarial examples.

**Scale-Invariant Nesterov Iterative Method (SI-NI-FGSM).** Lin et al. (2020) regards the adversarial example generation process as an optimization process, and propose two new attack methods, namely Nesterov Iterative Fast Gradient Sign Method (NI-FGSM) and Scale-Invariant attack Method (SIM), to improve the transferability of adversarial examples. In particular, the combination of SI-NI-M and TIM-DIM, namely SI-NI-TI-DIM, which greatly improves the success rate of black-box attack.

## 2.2. Defense Methods

In order to protect deep learning models from the threat of adversarial examples, various defense methods have been proposed to against them (Madry et al., 2018; Papernot et al., 2016; Xie et al, 2018; Guo et al., 2017; Samangouei et al, 2018; Mao et al., 2020;). Goodfellow et al. (2014) proposed to put the adversarial examples into the training data and get them involved in the model training process to improve the robustness of the model. Since ordinary adversarial training is still susceptible to adversarial examples, Tramer et al. (2018) suggested ensemble adversarial training to further enhance the robustness of the model, which means adversarial examples generated by multiple models separately are put into the training set of a single model to obtain a more robust classifier. Liu et al. (2019) proposed a JPEG-based defensive compression framework, namely "feature distillation", to effectively rectify adversarial examples without impacting classification accuracy on benign data. Cohen et al. (2019) use randomized smoothing to obtain an ImageNet classifier with certified adversarial robustness. Due to the error amplification effect of the standard denoiser, the small residual adversarial noise is gradually amplified, leading to the wrong classification of the model. For this reason, Liao et al. (2018) put forward high-level representation guided denoiser to purify the adversarial examples. Jia et al. (2019) utilize an end-to-end image compression model to defend adversarial examples. The above mentioned defense methods can be generally classified into two types: input modification and network structure or parameter modification. In this paper, we aim to generate more transferable adversarial examples to evaluate the model and improve its robustness.

## 3. Methodology

### 3.1. Motivation

Under the white-box setting, the adversarial examples tend to demonstrate strong attack capability, while in the black-box setting, the attack performance is poor, which we believe is the result of overfitting of adversarial examples, that is, the attack performance variance of the same adversarial examples in the white-box and black-box setting which is similar to the performance difference of the same neural network in the training set and the test set.

Analogous to the process of training deep learning models, the process of generating adversarial examples can also be viewed as an optimization problem. Therefore, we can apply the methods used to improve the generalization performance of deep learning models to the generation process of adversarial examples, so as to improve the transferability of adversarial examples.

Methods to improve the generalization performance of deep learning models are mainly as follows: (1) improve data quality and quantity; (2) use better optimization algorithm; (3) enhance optimization algorithm tuning; (4) use multiple models. Correspondingly, methods to improve the transferability of adversarial examples can also start from the following four aspects: (1) data augmentation, such as diverse inputs (Xie et al., 2019); (2) choose better optimization algorithms, such as Momentum Iterative Fast Gradient Sign Method (MI-FGSM) and Nesterov Iterative Fast Gradient Sign Method (NI-FGSM); (3) enhance optimization algorithm tuning, such as using multiple iterations and experiments to find better value of hyper-parameters; (4) model augmentation, such as attacking an ensemble of models mentioned in MI-FGSM, the transformation-invariant attack mentioned in TIM (Dong et al., 2019) and SI-NI-FGSM (Lin et al., 2020), and the integration of several attack methods which can also be considered as another form of model augmentation(Dong et al., 2019; Lin et al., 2020). Based on the above analysis, and considering the existing adversarial examples generation methods, we introduce AdaBelief optimizer and crop invariance into the adversarial example generation process, and propose AdaBelief Iterative Fast Gradient Method (ABI-FGM) and Crop-Invariant attack Method (CIM) to generate more transferable adversarial examples.

### 3.2. AdaBelief Iterative Fast Gradient Method

Adabelief (Zhuang et al., 2020) is an adaptive learning rate optimization algorithm, which can be easily modified from Adam (Kingma et al., 2020) without additional parameters. The intuition for AdaBelief is to adapt the step size according to the "belief" in the current gradient direction. The exponential moving average (EMA) of the noise gradient, i.e., $m_t$ is regarded as the prediction of the gradient at the next time step. If the observed gradient deviates greatly from the prediction, we will not trust the current observation and hence take a small step. However, if the observed gradient is close to the prediction, we believe it and take a large step to speed up the decline of the loss function in the dimensions with smaller gradient, to escape from poor local minimum, and to make loss function converge better. Adabelief optimizer takes convergence speed and generalization performance into consideration. Therefore, we apply the Adabelief optimizer to generate adversarial examples and obtain tremendous benefits from the perspective that the adversarial examples generation process is similar to the training process of deep neural networks. By applying the Adabelief optimizer to the process of generating adversarial examples, we propose the Adabelief Iterative Fast Gradient Method (ABI-FGM) to improve the transferability of adversarial examples.

The AdaBelief Iterative Fast Gradient Method (ABI-FGM) is summarized in Algorithm 1. Specifically speaking, it not only accumulates velocity vector along the gradient direction of the loss function in the iteration process, but also accumulates the square values of the difference between the predicted gradient and the gradient by weight. Then, after obtaining the direction of parameter updates based on two vectors, ABI-FGM adjusts the parameters needed to update with the purpose of adjusting the step size according to deviation between the actual gradient and predicted one, thus not only ensuring convergence speed, but also guaranteeing the convergence effect. As for the updating direction, the gradient of each iteration $\nabla_x J(\theta, x_t^*, y)$ should be normalized by its own $L_1$ distance, because the scale of these gradients differs widely in each iteration [8]. Similar to Adam (Kingma et al., 2020), $m_t$ accumulates the gradients of the first $t$ iterations with a decay factor $\beta_1$, defined in Eq. (6). $s_t$ accumulates the square values of the deviation between the gradient of the first iteration and $m_t$ with a decay factor $\beta_2$, defined in Eq. (7). The value of the hyper-parameter $\beta_1$ and $\beta_2$ is usually between (0,1). The update direction of $x$ is given by Eq. (9), where the stability factor $\delta$ is set to prevent the denominator from being zero. By doing this, the update on the dimensions with smaller gradient can be sped up. We adopt the $L_\infty$ norm bound, the limit requirement of which, unlike previous methods, is not met with the sign function of gradient, but with the corresponding step size and update direction within the corresponding $L_2$ norm bound. Then, adversarial examples are constrained within the $L_\infty$ norm bound using the $\text{clip}_x^\varepsilon(\cdot)$ function, just as what is defined in Eq. (9) and Eq. (10).

### 3.3. Crop-Invariant Attack Method

Inspired by data augmentation (Krizhevsky et al., 2017; Simonyan et al., 2014), we perform random cropping on images within a certain range, then calculate the loss functional values of the original image and the cropped image, and classify these images. The experimental results show that deep neural networks have crop-invariant property, which means on the same deep neural network, the loss function values of the original image and the cropped image are close to each other, and the classification accuracy rates of the original image and the cropped image are close to each other. (see Section 4.2 for detailed experimental results.) We think this is an intriguing property of the boundary area of an image; in other words, the most important part of an image tends to be in the very center, and the closer to the boundary a part is, the less important it is. This is also consistent with human habits. When we take pictures or display images, we tend to focus the most important part in the center position. Cropping the boundary area of the image can remove the less important parts of the image and realize the loss-preserving transformation of the image (Lin et al., 2020). Based on the aforementioned analysis, we propose a crop invariant attack method, which optimizes the adversarial perturbations over the crop copies of the input images:

$$\arg\max_{x^{adv}} \sum_{i=0}^{m-1} w_i J(\theta, C_i(x^{adv}), y), \quad (11)$$

$$\text{s.t.} \|x^{adv} - x\|_\infty \le \varepsilon, \sum_{i=0}^{m-1} w_i = 1, \quad (12)$$

**Algorithm 1** ABI-FGM

**Input:** A clean example $x$ with ground-truth label $y$; a classifier $f$ with loss function $J$;
**Input:** Perturbation size $\varepsilon$; maximum iterations $T$; AdaBelief decay factors $\beta_1$ and $\beta_2$; and a denominator stability factor $\delta$.
**Output:** An adversarial example $x^{adv}$
1: $\alpha = \varepsilon / T$; $x_0^{adv} = x$; $g_0 = 0$; $m_0 = 0$; $s_0 = 0$
2: **for** $t = 0$ to $T - 1$ **do**
3:  Get $g_{t+1}$ by $g_{t+1} = \dfrac{\nabla_x J(\theta, x_t^{adv}, y)}{\|\nabla_x J(\theta, x_t^{adv}, y)\|_1}$
4:  Update $m_{t+1}$ by $m_{t+1} = \beta_1 m_t + (1-\beta_1) g_{t+1}$ (6)
5:  Update $s_{t+1}$ by $s_{t+1} = \beta_2 \cdot s_t + (1-\beta_2) \cdot (g_{t+1} - m_{t+1})^2$ (7)
6:  Update $m_{t+1} = \dfrac{m_{t+1}}{1-\beta_1^{t+1}}$; $s_{t+1} = \dfrac{s_{t+1} + \delta}{1-\beta_2^{t+1}}$ (8)
7:  Update $x_{t+1}^{adv}$ by
$$x_{t+1}^{adv} = x_t^{adv} + \alpha \cdot \dfrac{m_{t+1}}{\sqrt{s_{t+1}} + \delta} \Big/ \left\|\dfrac{m_{t+1}}{\sqrt{s_{t+1}} + \delta}\right\|_2 \quad (9)$$
8:  Update $x_{t+1}^{adv}$ by $x_{t+1}^{adv} = \text{Clip}_x^\varepsilon(x_{t+1}^{adv})$ (10)
9: **return** $x^{adv} = x_T^{adv}$

where $C_i(x)$ is the crop function of the $i$-th crop copy of the input image $x$, $w_i$ is the corresponding weight, $m$ denotes the number of the crop copies. With this method, we can effectively achieve ensemble attacks on multiple models by model augmentation, so as to avoid the overfitting on white-box model being attacked and improves the transferability of adversarial examples.

### 3.4. Attack Algorithms

Moreover, CIM can also be readily combined with MI-FGSM and NI-FGSM, as CI-MI-FGSM and CI-NI-FGSM. The details in Appendix A. For the gradient-based generation process of adversarial examples, ABI-FGM introduces a better optimization algorithm to adaptively scales the step size and optimize the convergence process. For the ensemble attack of generating adversarial examples, CIM introduces model augmentation to derive multiple models to attack from a single model. CIM can be naturally combined with ABI-FGM to form a stronger attack, which we refer to as CI-AB-FGM (Crop-invariant Adabelief Iterative Fast Method). The algorithm of CI-AB-FGM attack is summarized in Algorithm 2 in Appendix A. Moreover, CIM can also be readily combined with MI-FGSM and NI-FGSM, as CI-MI-FGSM and CI-NI-FGSM. The CI-MI-FGSM and CI-NI-FGSM are summarized in Appendix A.

In addition, to further improve the transferability of adversarial examples, CI-AB-FGM can be combined with DIM (Diverse Input Method), TIM (Translation-Invariant Method), SI-NI-FGSM (Scale-Invariant Nesterov Iterative Fast Gradient Sign Method) and SI-NI-TI-DIM (Scale-Invariant Nesterov Iterative FGSM integrated with translation-invariant diverse input method) to form CI-AB-DIM, CI-AB-TIM, CI-AB-SIM and CI-AB-SI-TI-DIM respectively. The detailed algorithms for these attack methods are provided in Appendix A.

### 3.5. Relationships between Different Attacks

We sort out the relationship between our methods and other gradient-based attack methods in terms of better optimization algorithms and model augmentation, as shown in Figure 1. In summary:

—From the perspective of introducing better optimization algorithms, I-FGSM converts FGSM from single-step to multi-step iteration version. MI-FGSM introduces momentum term on the basis of I-FGSM, while NI-FGSM introduces Nesterov Accelerated Gradient on the basis of momentum method. Our method, ABI-FGM, is an improvement on I-FGSM by bringing the Adabelief optimizer into I-FGSM. These improvements optimize the convergence process so that more transferable adversarial examples can be generated.

—From the perspective of model augmentation, there is a connection between the crop-invariant attack method, other invariant attack methods and the fundamental momentum method. They can be connected through different conditional settings, as follows.

—If the input image is not going through loss-preserving transformation (Lin et al., 2020), that is, $C_i(x) = x$, CI-MI-FGSM is degraded to MI-FGSM.

—Through analysis, it is known that crop-invariant property is similar to translation-invariant property in a certain range. However, considering the complexity of the optimization equation of translation-invariant property, we need to expand the number of transformation, that is $C_i(x) \to C_{ij}(x)$, $w_i \to w_{ij}$, to make the number of cropping transformation and translation transformation done with each original image the same. On this basis, if the translation function is equal to the cropping function at each time, and the corresponding weight is equal, that is, $C_{ij}(x) = T_{ij}(x)$, $w_{ij}^C = w_{ij}^T$, then CI-MI-FGSM degrades to TI-MI-FGSM.

—if $x_t^{nes} = x_t^{adv} + \alpha \cdot \mu \cdot g_t$, $C_i(x) = S_i(x)$, $w_i = 1/m$, CI-MI-FGSM degrades to SI-NI-FGSM. In other words, the conversion between CI-MI-FGSM and SI-NI-FGSM requires not only the constraint of image transformation, but also the constraint exerted by Nesterov accelerated gradient.

In addition, our CI-MI-FGSM method can also be associated with other FGSM-related methods via different conditional settings. More importantly, clarifying the relationship between different methods will not only be beneficial for us to understand the existing attack methods, but also help us to explore and propose the stronger attack methods to generate more transferable adversarial examples.

## 4. Experiments

In this section, we intend to conduct extensive experiments to prove the efficiency and merits of the proposed methods. First, in section 4.1, we will specify the experiment settings; then, in section 4.2, we will explore the crop-invariance property for deep neural networks; in the following two sections, results in normally training model and adversarial training models will be compared using our methods and existing baseline methods respectively ; in section 4.5, the efficiency of our proposed methods will be further validated by the experimental results in other advanced defense models; last, discussion regarding our research ideas and future research directions will be presented in section 4.6.

### 4.1. Experimental Settings

**Dataset.** It is less meaningful to craft adversarial examples from the original images that are already classified wrongly. We randomly select 1000 images belonging to 1000 categories (i.e., one image per category) from the ImageNet verification set, which were correctly classified by our testing networks. All images were adjusted to 299×299×3.

**Networks.** We consider seven networks. The four normally trained networks are Inception-v3 (Inc-v3) (Szegedy et al., 2016), Inception-v4 (Inc-v4) (Szegedy et al., 2017), Inception-Resnet-v2 (IncRes-v2) (Szegedy et al., 2017), and Resnet-v2-101 (Res-101) (He et al., 2016); the three adversarially trained networks [23] (Tramr et al., 2018) are ens3-adv-Inception-v3 (Inc-v3$_{ens3}$), ens4-adv-Inception-v3 (Inc-v3$_{ens4}$), and ens-adv-Inception-ResNet-v2 (IncRes-v2$_{ens}$).

**Baselines.** We integrate our methods with DIM (Xie et al., 2019), TIM, and TI-DIM (Dong et al., 2019), SI-NI-FGSM and SI-NI-TI-DIM (Lin et al., 2020), to show the performance improvement of CI-AB-FGM over these baselines. We denote the attacks combined with our CI-AB-FGM as CI-AB-DIM, CI-AB-TIM, CI-AB-TI-DIM, CI-AB-SIM and CI-AB-SI-TI-DIM, respectively.

**Implementation details.** For the parameters of different attackers, we follow the default settings in (Dong et al., 2018) with the maximum perturbation $\varepsilon = 16$, number of iterations $T=10$, and step size $\alpha=1.6$. For MI-FGSM, the decay factor is defaulted to $\mu=1.0$. For DIM, we adopt the default transformation probability $p=0.5$. For TIM, we adopt the Gaussian kernel and the size of the kernel is set to 7×7. For SI-NI-FGSM, the number of scale copies is set to $m=5$, it should be noted that when our method is combined with SI-NI-FGSM, the scale factor in SIM is not set by default, but is randomly selected between [0.1,1] each time. For our CI-AB-FGM, the number of crop copies is set to $m=5$. In order to take into account the role of each transformation, the weight $w_i$ is the same every time, i.e., $w_i=1/5$. For crop function $C(\cdot)$, the input image $x$ is first randomly crop to a $rnd \times rnd \times 3$ image, with $rnd \in [279, 299)$, and then padded to the size 299×299×3 in a random manner. For intuitive understanding, Appendix B shows some images after random cropping and padding.

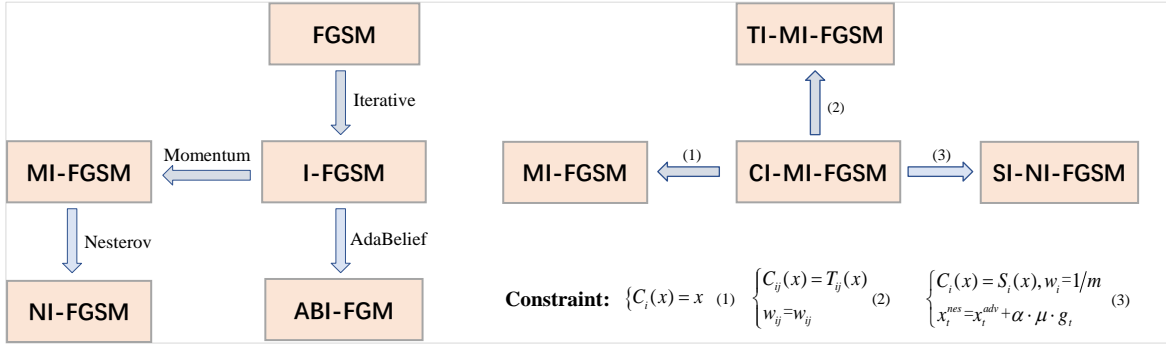

*Figure 1.* Relationships between different attacks

### 4.2. Crop-Invariant Property

In this section, we first verify the crop-invariant property of deep neural networks. we randomly select 1,000 original images from ImageNet dataset and keep the width of the cropped area randomly ranges from 0 to 40 with a step size 2 (i.e., the input image $x$ is randomly crop to a $rnd \times rnd \times 3$ image, with $rnd \in [299, 299]$ range to $rnd \in [259, 299)$ ). Then we input the crop images into the testing models, including Inc-v3, Inc-v4, IncRes-2, and Res-101, to get the average loss over 1,000 images.

As shown in Figure 2, it can be seen that the loss curves are generally stable when the width of the cropped area is in range [0,20]. That is, the loss values are very similar for the original and cropped images. Therefore, we regard that a cropped image is almost the same as the corresponding original image as inputs to the models, and we assume that the crop-invariant property of deep models is held within $rnd \in [279, 299)$ (i.e., the width of the cropped area is 20).

### 4.3. Attack a Single Model

In this section, we first compare the success rates of MI-FGSM, NI-FGSM, SI-NI-FGSM and our methods, then we integrate our CI-AB-FGM with DIM, TIM, TI-DIM, SI-NI-FGSM and SI-NI-TI-DIM respectively, and compare the black-box attack success rates of our extensions with the baselines under single model setting. We report the success rates of attacks in Table 1 for our methods and other three methods, Table 2 for DIM and CI-AB-DIM, Table 3 for TIM and CI-AB-TIIM, Table 4 for TI-DIM and CI-AB-TI-DIM, Table 5 for SI-NI-FGSM and CI-AB-SIM, Table 6 for SI-NI-TI-DIM and CI-AB-SI-TI-DIM.

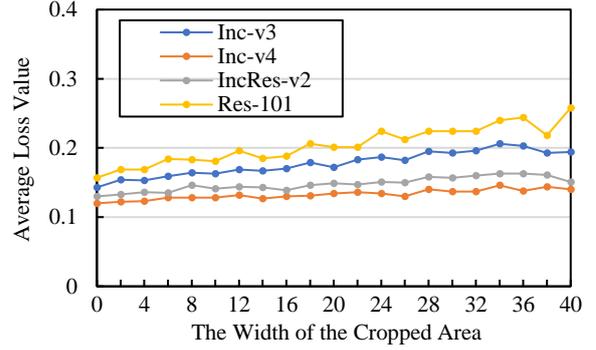

*Figure 2.* The average losses for Inc-v3, Inc-v4, IncRes-v2 and Res-101 at each crop width. The results are averaged over 1000 images.

As shown in Table 1, Our CI-AB-FGM has the highest attack success rate on the adversarially training networks under the black-box setting. In addition, as shown in Table 2-6, our extension methods consistently outperform the baseline attacks by 5% ~ 30% under the black-box setting, and achieve nearly 100% success rates under the white-box setting. It indicates that CI-AB-FGM can serve as a powerful method to boost the transferability of adversarial examples.

### 4.4. Attack an Ensemble of Models

As demonstrated in (Liu et al., 2016), attacking multiple models at the same time can improve the transferability of the generated adversarial examples. Therefore, we consider to show the performance of our methods by attacking multiple models simultaneously. Specifically, we attack the ensemble of Inc-v3, Inc-v4, IncRes-v2 and Res-101 with equal ensemble weights using DIM, CI-AB-DIM, TIM, CI-AB-TIM, TI-DIM, CI-AB-TI-DIM, SI-NI-FGSM, CI-AB-SIM, SI-NI-TI-DIM and CI-AB-SI-TI-DIM, respectively.

In Table 7, we show the results of ensemble-based attacks against the seven models. our methods improve the success rates across all experiments on challenging adversarially training networks over the baseline attacks. Especially, CI-AB-SI-TI-DIM, the extension by combining CI-AB-FGM with SI-NI-TI-DIM, can fool the adversarially trained models with a high average success rate of 95.3%, which outperforms the state-of-the-art gradient-based attack. The results in the paper demonstrate that these advanced adversarially trained models provide little robustness guarantee under the black-box attack of CI-AB-SI-TI-DIM.

## 5. Conclusions

In this paper, we propose AdaBelief Iterative Fast Gradient Method (ABI-FGM) and Crop-Invariant attack Method (CIM). Specifically speaking, ABI-FGM introduces the AdaBelief optimizer into the generating process of adversarial examples to optimize the convergence process. CIM optimizes the adversarial perturbations over the crop copies of input image to generate adversarial examples with more portability. ABI-FGM can be naturally combined with CIM to build up a strong adversarial attack, namely CI-AB-FGM.

*Table 1.* The success rates (%) of adversarial attacks against seven models under single model setting. The adversarial examples are crafted on Inc-v3 using MI-FGSM, NI-FGSM, SI-NI-FGSM and our methods. * indicates the white-box attacks.

| Attack | Inc-v3* | Inc-v4 | IncRes-v2 | Res-101 | Inc-v3$_{ens3}$ | Inc-v3$_{ens4}$ | IncRes-v2$_{ens}$ |
|---|---|---|---|---|---|---|---|
| MI-FGSM | **100.0** | 48.8 | 46.8 | 39.3 | 15.8 | 14.6 | 7.5 |
| NI-FGSM | **100.0** | 54.2 | 52.6 | 43.5 | 13.8 | 13.8 | 7.7 |
| ABI-FGM(**Ours**) | **100.0** | 50.1 | 47.3 | 40.5 | 18.0 | 17.3 | 8.1 |
| SI-NI-FGSM | **100.0** | 75.8 | 70.3 | 66.5 | 30.0 | 28.6 | 13.6 |
| CI-MI-FGSM(**Ours**) | 99.3 | **84.1** | 79.2 | **73.0** | 31.6 | 29.6 | 14.6 |
| CI-NI-FGSM(**Ours**) | 99.7 | 82.8 | 79.6 | 71.1 | 26.0 | 25.2 | 12.0 |
| CI-AB-FGM(**Ours**) | 99.6 | 83.6 | **80.2** | 70.2 | **35.8** | **33.2** | **17.6** |

*Table 2.* The success rates (%) of adversarial attacks against seven models under single model setting. The adversarial examples are crafted on Inc-v3, Inc-v4, IncRes-v2, and Res-152 respectively using DIM and CI-AB-DIM. * indicates the white-box attacks.

| Model | Attack | Inc-v3 | Inc-v4 | IncRes-v2 | Res-101 | Inc-v3ens3 | Inc-v3ens4 | IncRes-v2ens |
|---|---|---|---|---|---|---|---|---|
| **Inc-v3** | DIM | 99.1* | 69.9 | 64.1 | 59.2 | 22.5 | 21.1 | 10.1 |
| | CI-AB-DIM(**Ours**) | **99.3*** | **86.5** | **83.3** | **75.2** | **40.4** | **37.6** | **19.3** |
| **Inc-v4** | DIM | 86.5 | **99.5*** | 79.4 | 73.9 | 35.4 | 32.7 | 19.7 |
| | CI-AB-DIM(**Ours**) | **91.9** | **99.5*** | **86.2** | **77.3** | **43.9** | **41.9** | **25.9** |
| **IncRes-v2** | DIM | 86.3 | 85.1 | **99.5*** | 78.6 | 47.7 | 41.7 | 30.9 |
| | CI-AB-DIM(**Ours**) | **91.6** | **90.1** | 98.9* | **83.5** | **61.8** | **54.4** | **47.6** |
| **Res-101** | DIM | 80.8 | 74.7 | 73.3 | 98.0* | 43.7 | 38.4 | 24.9 |
| | CI-AB-DIM(**Ours**) | **90.9** | **85.9** | **87.4** | **98.4*** | **62.2** | **56.8** | **43.4** |

*Table 3.* The success rates (%) of adversarial attacks against seven models under single model setting. The adversarial examples are crafted on Inc-v3, Inc-v4, IncRes-v2, and Res-152 respectively using TIM and CI-AB-TIM. * indicates the white-box attacks.

| Model | Attack | Inc-v3 | Inc-v4 | IncRes-v2 | Res-101 | Inc-v3ens3 | Inc-v3ens4 | IncRes-v2ens |
|---|---|---|---|---|---|---|---|---|
| **Inc-v3** | TIM | **100.0*** | 51.8 | 47.4 | 41.3 | 25.7 | 22.3 | 14.4 |
| | CI-AB-TIM(**Ours**) | 99.4* | **82.9** | **76.8** | **67.1** | **54.0** | **49.7** | **36.3** |
| **Inc-v4** | TIM | 63.7 | **99.8*** | 54.9 | 46.1 | 31.2 | 28.1 | 18.9 |
| | CI-AB-TIM(**Ours**) | **87.1** | 99.0* | **79.4** | **67.9** | **56.1** | **51.5** | **42.2** |
| **IncRes-v2** | TIM | 71.4 | 65.5 | **99.6*** | 58.1 | 39.3 | 34.7 | 29.0 |
| | CI-AB-TIM(**Ours**) | **89.8** | **87.6** | 99.2* | **79.1** | **69.7** | **62.1** | **61.5** |
| **Res-101** | TIM | 54.1 | 46.8 | 47.3 | **98.3*** | 30.8 | 28.8 | 21.0 |
| | CI-AB-TIM(**Ours**) | **84.9** | **80.9** | **79.8** | **98.3*** | **68.7** | **62.2** | **55.3** |

*Table 4.* The success rates (%) of adversarial attacks against seven models under single model setting. The adversarial examples are crafted on Inc-v3, Inc-v4, IncRes-v2, and Res-152 respectively using TI-DIM and CI-AB-TI-DIM. * indicates the white-box attacks.

| Model | Attack | Inc-v3 | Inc-v4 | IncRes-v2 | Res-101 | Inc-v3ens3 | Inc-v3ens4 | IncRes-v2ens |
|---|---|---|---|---|---|---|---|---|
| **Inc-v3** | TI-DIM | **99.5*** | 79.1 | 74.2 | 68.9 | 49.2 | 45.8 | 29.8 |
| | CI-AB-TI-DIM(**Ours**) | 98.7* | **83.8** | **80.5** | **72.7** | **60.3** | **56.6** | **42.2** |
| **Inc-v4** | TI-DIM | 84.4 | **99.4*** | 77.0 | 68.8 | 51.1 | 48.8 | 36.8 |
| | CI-AB-TI-DIM(**Ours**) | **89.6** | 98.9* | **83.2** | **74.3** | **61.8** | **59.1** | **49.0** |
| **IncRes-v2** | TI-DIM | 87.8 | 85.6 | **99.4*** | 78.8 | 64.8 | 57.8 | 55.5 |
| | CI-AB-TI-DIM(**Ours**) | **91.0** | **90.3** | 98.9* | **82.7** | **75.1** | **68.3** | **68.0** |
| **Res-101** | TI-DIM | 78.0 | 72.4 | 72.2 | **98.1*** | 57.0 | 54.6 | 43.9 |
| | CI-AB-TI-DIM(**Ours**) | **87.7** | **84.7** | **85.5** | 97.8* | **73.7** | **71.8** | **61.5** |

Table 5. The success rates (%) of adversarial attacks against seven models under single model setting. The adversarial examples are crafted on Inc-v3, Inc-v4, IncRes-v2, and Res-152 respectively using SI-NI-FGSM and CI-AB-SIM. * indicates the white-box attacks.

| Model | Attack | Inc-v3 | Inc-v4 | IncRes-v2 | Res-101 | Inc-v3ens3 | Inc-v3ens4 | IncRes-v2ens |
|---|---|---|---|---|---|---|---|---|
| **Inc-v3** | SI-NI-FGSM | **100.0*** | 75.8 | 70.3 | 66.5 | 30.0 | 28.6 | 13.6 |
| | CI-AB-SIM(**Ours**) | 98.1* | **88.8** | **85.8** | **82.3** | **57.1** | **52.8** | **34.4** |
| **Inc-v4** | SI-NI-FGSM | 85.7 | **98.9*** | 80.9 | 73.9 | 43.3 | 39.2 | 22.7 |
| | CI-AB-SIM(**Ours**) | **92.1** | 98.7* | **88.8** | **82.2** | **65.4** | **62.3** | **47.3** |
| **IncRes-v2** | SI-NI-FGSM | 82.5 | 76.7 | 95.8* | 70.9 | 43.8 | 37.1 | 28.7 |
| | CI-AB-SIM(**Ours**) | **92.3** | **91** | **97.5*** | **86.1** | **75.6** | **69.5** | **65.1** |
| **Res-101** | SI-NI-FGSM | 72.9 | 64.9 | 64.4 | 96.1 | 33.6 | 29.6 | 17.8 |
| | CI-AB-SIM(**Ours**) | **86.6** | **84.7** | **83.3** | **97.6*** | **65.7** | **60.3** | **50.1** |

Table 6. The success rates (%) of adversarial attacks against seven models under single model setting. The adversarial examples are crafted on Inc-v3, Inc-v4, IncRes-v2, and Res-152 respectively using SI-NI-TI-DIM and CI-AB-SI-TI-DIM. * indicates the white-box attacks.

| Model | Attack | Inc-v3 | Inc-v4 | IncRes-v2 | Res-101 | Inc-v3ens3 | Inc-v3ens4 | IncRes-v2ens |
|---|---|---|---|---|---|---|---|---|
| **Inc-v3** | SI-NI-TI-DIM | **99.1*** | 81.8 | 77.8 | 70.4 | 54.8 | 52.7 | 37.9 |
| | CI-AB-SI-TI-DIM(**Ours**) | 97.9* | **88.3** | **84.3** | **79.4** | **73.5** | **70.9** | **56.6** |
| **Inc-v4** | SI-NI-TI-DIM | 86.4 | **99.3*** | 78.8 | 72.8 | 61.3 | 58.2 | 47.3 |
| | CI-AB-SI-TI-DIM(**Ours**) | **91.9** | 98.3* | **87.8** | **80.4** | **76.6** | **74.0** | **64.4** |
| **IncRes-v2** | SI-NI-TI-DIM | 84.8 | 84.3 | **98.9*** | 78.5 | 67.2 | 61.9 | 57.3 |
| | CI-AB-SI-TI-DIM(**Ours**) | **91.2** | **90.0** | 95.9* | **84.8** | **83.1** | **78.0** | **78.5** |
| **Res-101** | SI-NI-TI-DIM | 83.1 | 79.0 | 79.9 | **98.3*** | 67.3 | 63.5 | 52.4 |
| | CI-AB-SI-TI-DIM(**Ours**) | **86.5** | **84.1** | **81.5** | 97.7* | **78.0** | **75.5** | **68.2** |

Table 7. The success rates (%) of adversarial attacks against seven models under multi-model setting. The adversarial examples are crafted for the ensemble of Inc-v3, Inc-v4, IncRes-v2, and Res-152. * indicates the white-box models being attacked.

| Attack | Inc-v3* | Inc-v4* | IncRes-v2* | Res-101* | Inc-v3ens3 | Inc-v3ens4 | IncRes-v2ens | Average |
|---|---|---|---|---|---|---|---|---|
| DIM | **99.6** | **99.0** | 97.8 | **97.8** | 64.6 | 58.6 | 40.1 | 79.6 |
| CI-AB-DIM(**Ours**) | 99.0 | **99.0** | 97.9 | 97.5 | **74.3** | **69.0** | **53.5** | **84.3** |
| TIM | 99.9 | 99.5 | 99.1 | 99.5 | 68.9 | 63.7 | 53.3 | 83.4 |
| CI-AB-TIM(**Ours**) | 98.9 | 98.9 | 98.1 | 97.5 | **86.5** | **85.1** | **80.0** | **92.1** |
| TI-DIM | **99.0** | 98.2 | 97.5 | 97.4 | 81.7 | 77.5 | 70.4 | 88.8 |
| CI-AB-TI-DIM(**Ours**) | **99.0** | **98.3** | **98.0** | 97.4 | **87.8** | **86.4** | **82.1** | **92.7** |
| SI-NI-FGSM | 100.0 | 100.0 | 100.0 | 99.9 | 79.5 | 74.6 | 58.5 | 87.5 |
| CI-AB-SIM(**Ours**) | 99.6 | 99.6 | 99.5 | 99.5 | **93.9** | **92.3** | **84.2** | **95.6** |
| SI-NI-TI-DIM | 99.9 | 99.8 | 99.8 | 99.9 | 93.6 | 91.3 | 85.9 | 95.7 |
| CI-AB-SI-TI-DIM(**Ours**) | 99.4 | 99.0 | 98.9 | 98.3 | **96.1** | **96.1** | **93.6** | **97.3** |

Compared with current FGSM-related methods, extensive experiments on the ImageNet dataset show that the success rates of our methods under the black-box setting have improved considerably. Particularly, by combining our methods with the existing gradient-based attack ones, a stronger attack method is formed to further boost the black-box attack success rates. In addition, we also attack an ensemble of methods to increase the transferability of adversarial examples, which can be shown in our experimental results of the average success rate in attacking seven models being as high as 97.3% under the ensemble model setting of our best method CI-AB-SI-TI-DIM. Furthermore, we also attack advanced defense models to verify the effectiveness of our methods, our experimental results in the latest robustness defense methods indicate there is a higher attack success rate of our methods than of the current gradient-based attack methods. Our work in CI-AB-FGM demonstrates other methods used for enhancing the generalization performance of deep neural networks are also likely to attribute to strong attack, which is one of our future research areas, but is contingent on how to find effective methods and apply them to the generating process of adversarial examples. We hope that the proposed attack method can help evaluate the robustness of models and effectiveness of defense methods.

# Appendix A. Details of Algorithms

The algorithm of CI-AB-FGM, CI-NI-FGSM and CI-AB-SI-TI-DIM attacks is summarized in Algorithm 2, 3 and 4, respectively. We can obtain the CI-MI-FGSM attack algorithm by removing step 4 of Algorithm 3 and obtain the NI-FGSM attack algorithm by removing $C(\cdot)$ in step 6. In addition, we can get the CI-AB-TI-DIM attack algorithm by removing $S(\cdot)$ in Step 4 of Algorithm 4, and get the CI-AB-SIM attack algorithm by removing $T(\cdot;p)$ in Step 4 and step 6 of Algorithm 4. Of course, our method can also be related to the family of Fast Gradient Sign Methods via different parameter settings.

---

**Algorithm 2** CI-AB-FGM

**Input:** A clean example $x$ with ground-truth label $y$; a classifier $f$ with loss function $J$;
**Input:** Perturbation size $\varepsilon$; maximum iterations $T$; AdaBelief decay factors $\beta_1$ and $\beta_2$; and a denominator stability factor $\delta$.
**Output:** An adversarial example $x^{adv}$
1: $\alpha = \varepsilon/T$; $x_0^{adv} = x$; $g_0 = 0$; $m_0 = 0$; $v_0 = 0$
2: **for** $t = 0$ to $T-1$ **do**
3:   **for** $i = 0$ to $m-1$ **do**
4:     Get the gradients by $\nabla_x J(\theta, C_i(x_t^{adv}), y)$
5:   Get average gradients as
$$g_{t+1} = \sum_{i=0}^{m-1} w_i \nabla_x J(\theta, C_i(x_t^{adv}), y),$$
6:   Get $g_{t+1}$ by $g_{t+1} = \dfrac{g_{t+1}}{\|g_{t+1}\|_1}$
7:   Update $m_{t+1}$ by $m_{t+1} = \beta_1 m_t + (1-\beta_1) g_{t+1}$
8:   Update $s_{t+1}$ b
$$s_{t+1} = \beta_2 \cdot s_t + (1-\beta_2) \cdot (g_{t+1} - m_{t+1})^2$$
9:   Update $m_{t+1} = \dfrac{m_{t+1}}{1-\beta_1^{t+1}}$; $s_{t+1} = \dfrac{s_{t+1}+\varepsilon}{1-\beta_2^{t+1}}$
10:   Update $x_{t+1}^{adv}$ by
$$x_{t+1}^{adv} = x_t^{adv} + \alpha \cdot \dfrac{m_{t+1}}{\sqrt{s_{t+1}}+\varepsilon} \bigg/ \left\|\dfrac{m_{t+1}}{\sqrt{s_{t+1}}+\varepsilon}\right\|_2$$
11:   Update $x_{t+1}^{adv}$ by Eq. (5)
12: **return** $x^{adv} = x_T^{adv}$

---

# Appendix B. Visualization of images

The six randomly selected original images and corresponding randomly transformed images and generated adversarial samples are shown in Figure 3. The adversarial examples are crafted on Inc-v3 by the CI-AB-SI-TI-DIM method. We can see that these crafted adversarial perturbations are human imperceptible.

---

**Algorithm 3** CI-NI-FGSM

**Input:** A clean example $x$ with ground-truth label $y$; a classifier $f$ with loss function $J$;
**Input:** Perturbation size $\varepsilon$; maximum iterations $T$ and decay factor $\mu$; number of scale copies $m$.
**Output:** An adversarial example $x^{adv}$
1: $\alpha = \varepsilon/T$
2: $x_0^{adv} = x$; $g_0 = 0$
3: **for** $t = 0$ to $T-1$ **do**
4:   Get $x_t^{nes}$ by $x_t^{nes} = x_t^{adv} + \alpha \cdot \mu \cdot g_t$ ▷ make a jump in the direction of previous accumulated gradients
5:   **for** $i = 0$ to $m-1$ **do**
6:     Get the gradients by $\nabla_x J(\theta, C_i(x_t^{nes}), y)$
7:   Get average gradients as
$$g = \sum_{i=0}^{m-1} w_i \nabla_x J(\theta, C_i(x_t^{nes}), y),$$
8:   Update $g_{t+1}$ by $g_{t+1} = \mu \cdot g_t + \dfrac{g}{\|g\|_1}$
9:   Update $x_{t+1}^{adv}$ by Eq. (5)
10: **return** $x^{adv} = x_T^{adv}$

---

**Algorithm 4** CI-AB-SI-TI-DIM

**Input:** A clean example $x$ with ground-truth label $y$; a classifier $f$ with loss function $J$;
**Input:** Perturbation size $\varepsilon$; maximum iterations $T$; AdaBelief decay factors $\beta_1$ and $\beta_2$; and a denominator stability factor $\delta$.
**Output:** An adversarial example $x^{adv}$
1: $\alpha = \varepsilon/T$; $x_0^{adv} = x$; $g_0 = 0$; $m_0 = 0$; $v_0 = 0$
2: **for** $t = 0$ to $T-1$ **do**
3:   **for** $i = 0$ to $m-1$ **do**
4:     Get the gradients by
$$\nabla_x J(\theta, T(C(S_i(x_t^{adv})); p), y)$$
5:   Get average gradients as
$$g_{t+1} = \sum_{i=0}^{m-1} w_i \nabla_x J(\theta, T(C(S_i(x_t^{adv})); p), y),$$
6:   Convolve the gradients by $g_{t+1} = W * g_{t+1}$ ▷ convolve gradient with the pre-defined kernel $W$
7:   Get $g_{t+1}$ by $g_{t+1} = \dfrac{g_{t+1}}{\|g_{t+1}\|_1}$
8:   Update $m_{t+1}$ by $m_{t+1} = \beta_1 m_t + (1-\beta_1) g_{t+1}$
9:   Update $s_{t+1}$ by
$$s_{t+1} = \beta_2 \cdot s_t + (1-\beta_2) \cdot (g_{t+1} - m_{t+1})^2$$
10:   Update $m_{t+1} = \dfrac{m_{t+1}}{1-\beta_1^{t+1}}$; $s_{t+1} = \dfrac{s_{t+1}+\varepsilon}{1-\beta_2^{t+1}}$
11:   Update
$$x_{t+1}^{adv} = x_t^{adv} + \alpha \cdot \dfrac{m_{t+1}}{\sqrt{s_{t+1}}+\varepsilon} \bigg/ \left\|\dfrac{m_{t+1}}{\sqrt{s_{t+1}}+\varepsilon}\right\|_2$$
12:   Update $x_{t+1}^{adv}$ by Eq. (5)
13: **return** $x^{adv} = x_T^{adv}$

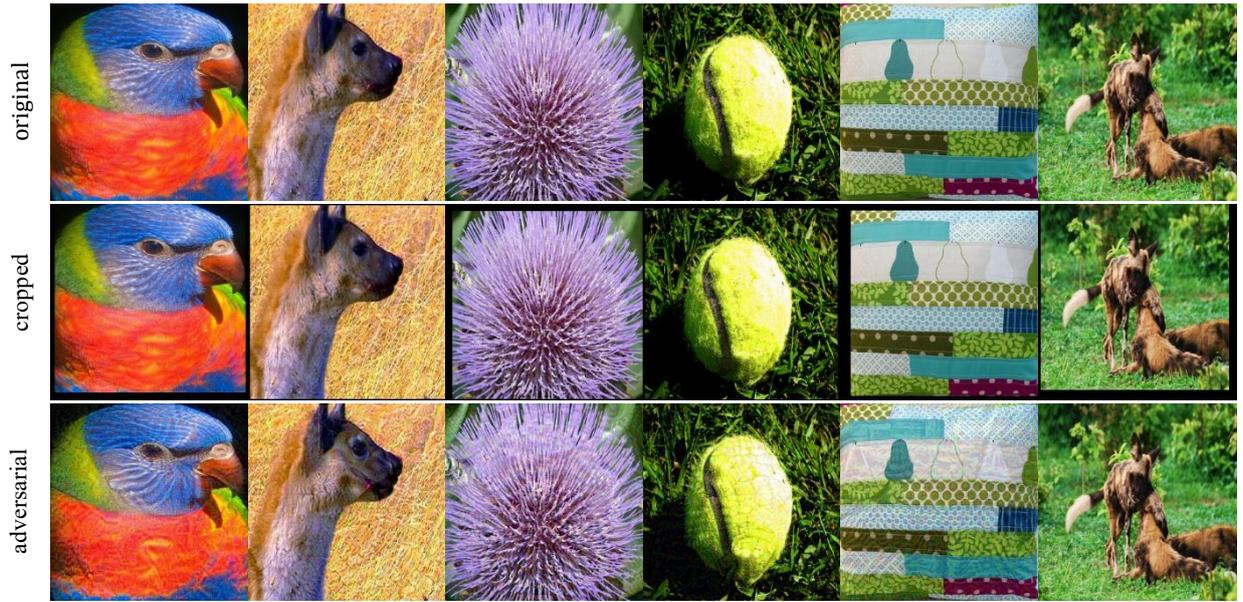

*Figure 3.* Images from first to third line are original inputs, randomly cropped images, and generated adversarial examples, respectively.